\def\localexplanation{
\foreach \inputid in {61, 110, 84}{
\centerline{
\subfloat{\includegraphics[width=.12\textwidth]{code_len_16_local/input_\inputid}}
\foreach \classid in {0,...,10}{
\subfloat{\includegraphics[width=.12\textwidth]{code_len_16_local/cifar_test_\inputid_441687_MNIST10_code_len=16_wl=0_\classid}}
}}}}

\def\localexplanationinf{
\foreach \inputid in {61, 110, 84}{
\centerline{
\subfloat{\includegraphics[width=.12\textwidth]{code_len_inf_local/input_\inputid}}
\foreach \classid in {0,...,10}{
\subfloat{\includegraphics[width=.12\textwidth]{code_len_inf_local/cifar_test_\inputid_5645154_MNIST10_code_len=256_\classid}}
}}}}

\def\imageswithoutrouters{
\foreach \expertid in {0,...,15}{
\centerline{
\subfloat{\frame{\includegraphics[width=1.25\textwidth,height=.06\textheight]{code_len_4_only_IC/\expertid}
}}
\\[-2ex]
}}}

\def\imageswithrouters{
\foreach \expertid in {0,...,15}{
\centerline{
\subfloat{\frame{\includegraphics[width=1.2\textwidth,height=.06\textheight]{code_len_4_all_losses/\expertid}
}}
\\[-2ex]
}}}
\documentclass{article}

\usepackage[preprint, nonatbib]{neurips_2021}

\usepackage[utf8]{inputenc} % allow utf-8 input
\usepackage[T1]{fontenc}    % use 8-bit T1 fonts
\usepackage{hyperref}       % hyperlinks
\usepackage{url}            % simple URL typesetting
\usepackage{booktabs}       % professional-quality tables
\usepackage{amsfonts}       % blackboard math symbols
\usepackage{nicefrac}       % compact symbols for 1/2, etc.
\usepackage{microtype}      % microtypography
\usepackage{xcolor}         % colors
\usepackage{amsmath}
\usepackage{graphicx}
\usepackage{subfig}
\usepackage{pgffor}
\usepackage{array}
\graphicspath{ {./images/} }
\title{Implicit Mixture of Interpretable Experts for Global and Local Interpretability}

\author{%
  Nathan Elazar\\
  Australian National University\\
  \texttt{nathan.elazar@anu.edu.au} \\
  \And 
  Kerry Taylor\\
  Australian National University\\
  \texttt{kerry.taylor@anu.edu.au}\\
}

\begin{document}

\maketitle

\begin{abstract}
  We investigate the feasibility of using mixtures of interpretable experts (MoIE) to build interpretable image classifiers on MNIST10. MoIE uses a black-box router to assign each input to one of many inherently interpretable experts, thereby providing insight into why a particular classification decision was made. We find that a naively trained  MoIE will learn to 'cheat', whereby the black-box router will solve the classification problem by itself, with each expert simply learning a constant function for one particular class. We propose to solve this problem by introducing interpretable routers and training the black-box router's decisions to match the interpretable router. In addition, we propose a novel implicit parameterization scheme that allows us to build mixtures of arbitrary numbers of experts, allowing us to study how classification performance, local and global interpretabillity vary as the number of experts is increased. Our new model, dubbed Implicit Mixture of Interpretable Experts (IMoIE) can match state-of-the-art classification accuracy on MNIST10 while providing local interpretabillity, and can provide global interpretabillity albeit at the cost of reduced classification accuracy.
\end{abstract}

\section{Introduction}
The field of Explainable AI (XAI) seeks to create human interpretable machine learning models. Typically, machine learning models are complicated and opaque, making it difficult for humans to understand them. Interpretable models can provide some explanation about their inner functioning, allowing for increased trust, transparency, and debuggability  \cite{xai_survey1} \cite{xai_survey2} \cite{xai_survey3}. Interpretability can be classified into 2 types: local and global. Local interpretability means that the model's prediction of a single input point can be understood, while global interpretability means that the model's behaviour over the entire input space can be understood. Most existing work in the XAI literature focuses on local interpretability, especially for high dimensional data, with global interpretability methods significantly neglected \cite{xai_survey4}. Global interpretability is of significant importance for debugging models, where local interpretability is often insufficient to understand and fix overall model behaviour.

While mixture of experts (MoE) models have been used extensively in deep learning for improving performance in terms of parameter efficiency and inference speed \cite{scaling_moe}, recently \cite{MoIE} proposed to use the MoE setup for the purpose of creating interpretable models. In the proposed mixture of interpretable experts (MoIE) framework, a black-box router sends inputs to be classified by one of many interpretable models. Specifically, \cite{MoIE} uses a deep neural network router which outputs a one-hot vector indicating which of many linear functions to use on a given input. \cite{MoIE} shows that this setup can match the performance of a black-box classifier, and claims that their model is interpretable. We first seek to verify these claims by reproducing this work on the MNIST10 image classification benchmark \cite{MNIST10}, where the learned experts can be visually inspected easily to verify that the experts are actually making meaningful decisions.

In the course of adapting this methodology to image classification we additionally make the following contributions:
\begin{itemize}
\item In order to prevent the black-box router solving the problem by itself, we augment each expert with its own interpretable router, and introduce a loss term to encourage the black-box router's output to match the interpretable router's decision. This encourages the black-box router's decisions to be as interpretable as possible. We demonstrate that this loss term is necessary for the experts to learn meaningful and interpretable functions.
\item Whereas \cite{MoIE} explicitly parameterizes all of their experts as linear functions, we \emph{implicitly} parameterize our experts by training a deconvolutional neural network to map a binary code indicating the ID of an expert to a linear function. This implicit parameterisation causes the experts to be naturally diverse and equally utilised, removing the need for the numerous auxiliary diversity loss terms used in \cite{MoIE}. In addition, this implicit parameterisation allows us to arbitrarily increase the number of experts used at essentially no additional computational cost.
\item We explore how the number of experts used impacts the global interpretability of the model. We propose a general framework for understanding interpretability, in which the number of experts used corresponds to an interpolation between global and local interpretability. When the number of experts used is small, they can all be inspected and the model's global behaviour understood. As more experts are added it becomes infeasible to inspect all of them, but individual experts can still be inspected to understand the model's behaviour for the small input region where that expert is active. In the limit of infinitely many experts, each expert is active for a single infinitesimal input point, thereby providing only local interpretability.
\end{itemize}

\section{Related Work}
\textbf{Post-hoc local interpretations} are the most popular methods for explaining image classifiers. These methods, such as \cite{phl1}\cite{phl2}\cite{phl3}, can be applied to any trained neural network architecture to provide local explanations of a single input prediction in the form of feature importance maps.

\textbf{Inherently locally interpretable models} have also been proposed for image classification. These methods, such as \cite{bcos}, design specialised neural network architectures which can provide local explanations along with their predictions.

\textbf{Globally interpretable models} have been studied much less extensively. As far as we are aware, all globally interpretable models rely on a black-box model for at least part of the decision making process. For example, \cite{protoptrees} use a deep neural network to learn a complicated distance function, and then classify new inputs by comparing distances to learned prototypes. While this model can explain which prototype was matched, the actual mechanics of why that prototype was matched is opaque. Similarly \cite{tcav} rely on a black-box neural network to judge how relevant a user defined concept is to a particular input. In contrast, our proposed method can produce an entirely interpretable model for global understanding.

\section{Methodology}
\subsection{Basic Implicitly parameterised Mixture of Experts}
Our proposed Implicitly parameterised Mixture of Experts (IMoE) model consists of an encoder network and a decoder network. The encoder maps an input sample $x \in \mathbb{R}^I$ to a vector of length $code\_length$ which is then binarized to create a code vector $c\in \{-1,1\}^{code\_length}$ by setting non-negative components to 1 and negative components to -1. Since each component of the code vector is discrete, we can interpret the encoder network as routing each input to one of $n=2^{code\_length}$ different experts. The decoder network then maps the code vector to an expert's parameters. Every expert is a linear function $E : \mathbb{R}^I \rightarrow \mathbb{R}^{O}$ where O is the number of output classes, and is used to classify the input sample. The output is given by 
$$\text{IMoIE}(x) = E(x) $$
where $E = \text{Decoder}(\text{Binarize}(\text{Encoder}(x)))$. 
Given a class label $y$, this entire model can be trained end-to-end to minimize the classification loss $L_{\text{C}}=\text{CrossEntropy}(\text{IMoIE}(x),y)$, utilising the straight-through estimator from \cite{ste} to backprop through $\text{Binarize}$. However, our experiments demonstrate that this basic model will learn to 'cheat' by having the encoder network solve the classification problem itself, and as a result each expert will learn a constant function which always outputs a single class.

\begin{figure}[t]
\caption{Diagram of IMoIE interpretable router training. The router $R$ for an input $x$ is generated by applying the Encoder and DecoderRouter. $R$ is then trained as a linear binary classifier to output 1 when applied to $x$ and $0$ when applied to another random sample $x^\prime$ which produced a different code. $@$ indicates matrix multiplication.}
\centerline{
\subfloat{\includegraphics[width=1.3\textwidth]{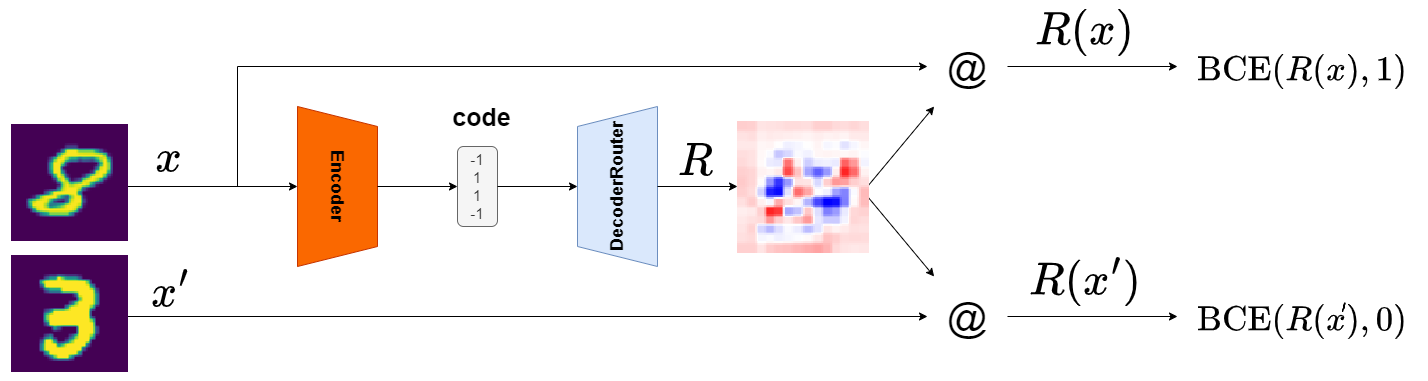}}}
\label{fig:imoiediagram}
\end{figure}

\subsection{Implicit Mixture of Interpretable Experts}
Our proposed Implicit Mixture of Interpretable Experts (IMOIE) is an IMOE where each expert is additionally equipped with an interpretable router. An IMOIE has two decoder networks, one which maps codes to experts (as before) and another which maps codes to routers. Each router is a linear function $R : \mathbb{R}^I \rightarrow \mathbb{R}$. The goal of each router is to perform the same job as the black-box encoder: the router should output $1$ for all inputs for which the black-box encoder outputs this router's own code, and $0$ for all other inputs. Formally, we introduce an additional routing loss term

$$L_{\text{R}}=\text{BinaryCrossEntropy}(R_i(x_i), 1) + \text{BinaryCrossEntropy}(R_i(x_{i+1}), \text{Indicator}(c_i == c_{i+1}))$$

where $R_i = \text{DecoderRouter}(\text{Binarize}(\text{Encoder}(x_i)))$ is the routing function selected for the $i$'th input data point in the batch, and the $\text{Indicator}$ function returns 1 when its argument is true or else 0. The routing loss term trains routers to output 1 for the input which produced their own code, and 0 for another random input that produced a different code. See Figure \ref{fig:imoiediagram} for diagram. By training the black-box encoder and the interpretable routers to agree on their outputs, we prevent the black-box encoder from solving the classification task by itself.

In addition, the IMOIE can be converted to a fully interpretable model by using the interpretable routers to make routing decisions, instead of the black-box encoder. This is achieved by evaluating the decoders on every possible code value, storing all of the resulting routers and experts in  memory, and then weighting all expert predictions by their routing scores. We refer to this as explicit mode IMoIE, with predictions given by
$$\text{Explicit\_IMoIE}(x) = \sum_{i=1}^{n} \text{Softmax}(R(x))_i E_i(x) $$
To further simplify the explicit model, we only take Softmax over the $8$ largest routing scores in $R(x)$, as we found that $8$ was the smallest number of scores we could use without significantly degrading performance. This is implemented by setting all but the $8$ largest components in $R(x)$ to $-\infty$ before Softmax is applied.
We also find it helpful to train the IMoIE to maximize its classification performance when in explicit mode, so we introduce another explicit classification loss term

$$L_{\text{EC}}= \text{CrossEntropy}(\text{Explicit\_IMoIE}(x), y)$$

For large code sizes it becomes computationally infeasible to materialize all experts and routers for every training step, so we instead only evaluate explicit mode using a random sample of experts and routers. In practice, we store the codes generated from the previous batch, and only materialize experts and routers from the previous batch of codes during training. We materialize all $2^{code\_length}$ experts and routers when evaluating explicit mode test performance.

We also use an $l_1$ regularizer on the experts and routers to encourage them to learn sparse weights.

$$L_1 = mean(|E|) + mean(|R|)$$

In summary, we train the model using

$$
Loss = L_{\text{C}}\lambda_{\text{C}} +
L_{\text{R}}\lambda_{R} + 
L_{\text{EC}}\lambda_{\text{EC}} +
L_1\lambda_1$$

\subsection{Implementation Details}
For MNIST10 \cite{MNIST10} we implement the encoder as a ResNet18 \cite{resnet} convolutional neural network which outputs a vector of length $code\_length$. The decoder is analogous to the encoder but with bilinear upsampling instead of maxpooling. The decoder outputs a feature map the same shape as the input but with 11 channels, and we treat this as a weight matrix defining linear functions for both the experts' output for each of the 10 classes and the router's single output. The feature map is reshaped and matrix multiplied with the input $x$.

We rescale image inputs from [0,1] to [-1,1]. We train for 100 epochs with a batch size of 128. We use the AdamW \cite{adam} optimizer with a learning rate of $3\times 10^{-4}$.

Unless otherwise specified, we use $\lambda_{\text{C}}=0.5$, $\lambda_{\text{R}}=0.7$, $\lambda_{\text{EC}}=1$, $\lambda_1=0.1$

\section{Experiments}

\subsection{Effect of Training Interpretable Routers on Learned Experts}

\begin{figure}[!ht]
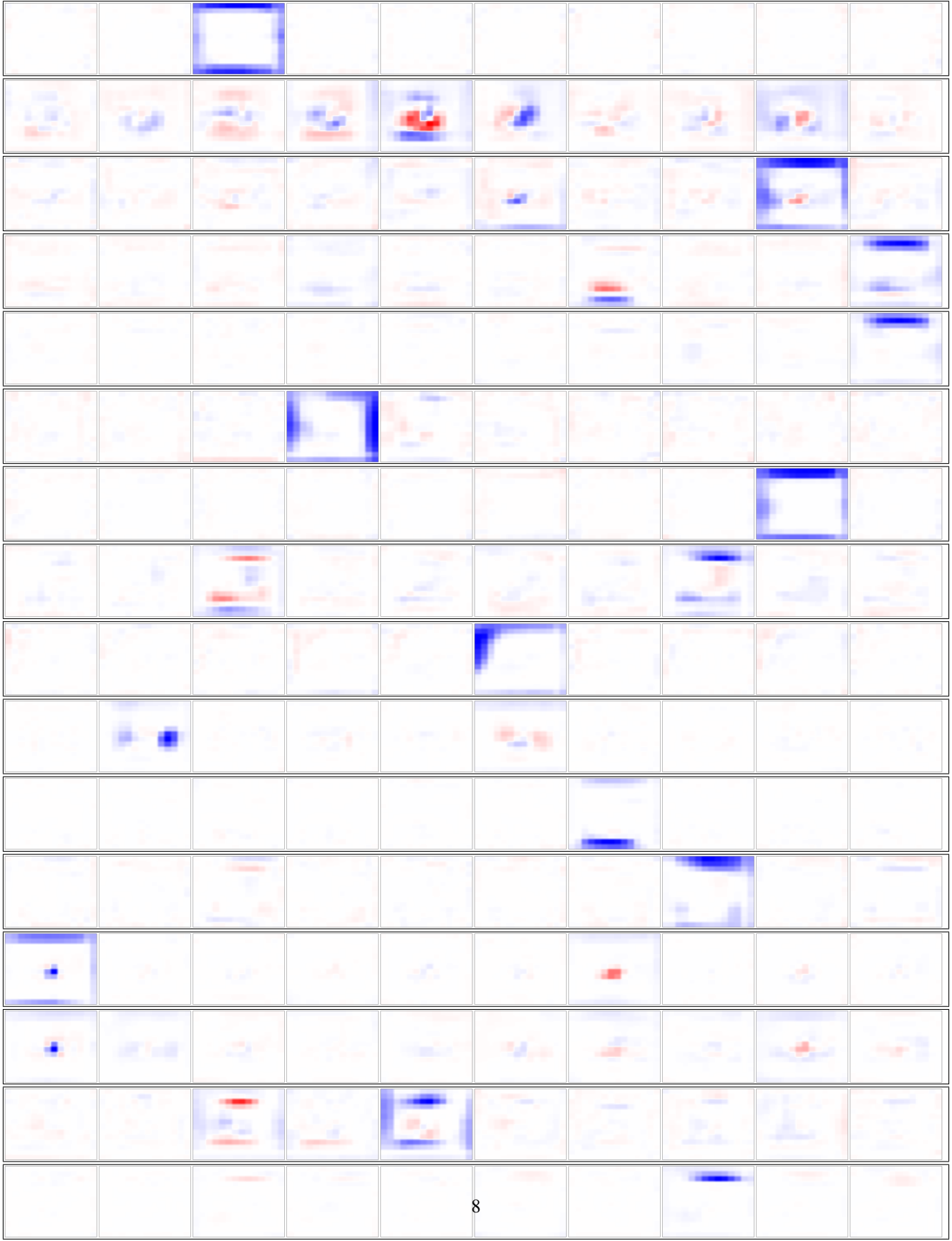

% begin images without
    \caption{All linear weights of IMoIE with $code\_length=4$ trained using only $L_{\text{C}}$ and $L_1$. Each row corresponds to the linear function of one expert, with columns corresponding to classes 0 through 9. Red is positive, blue is negative}
    \label{fig:full_without_reject}
\imageswithoutrouters
\end{figure}

\begin{figure}[!ht]
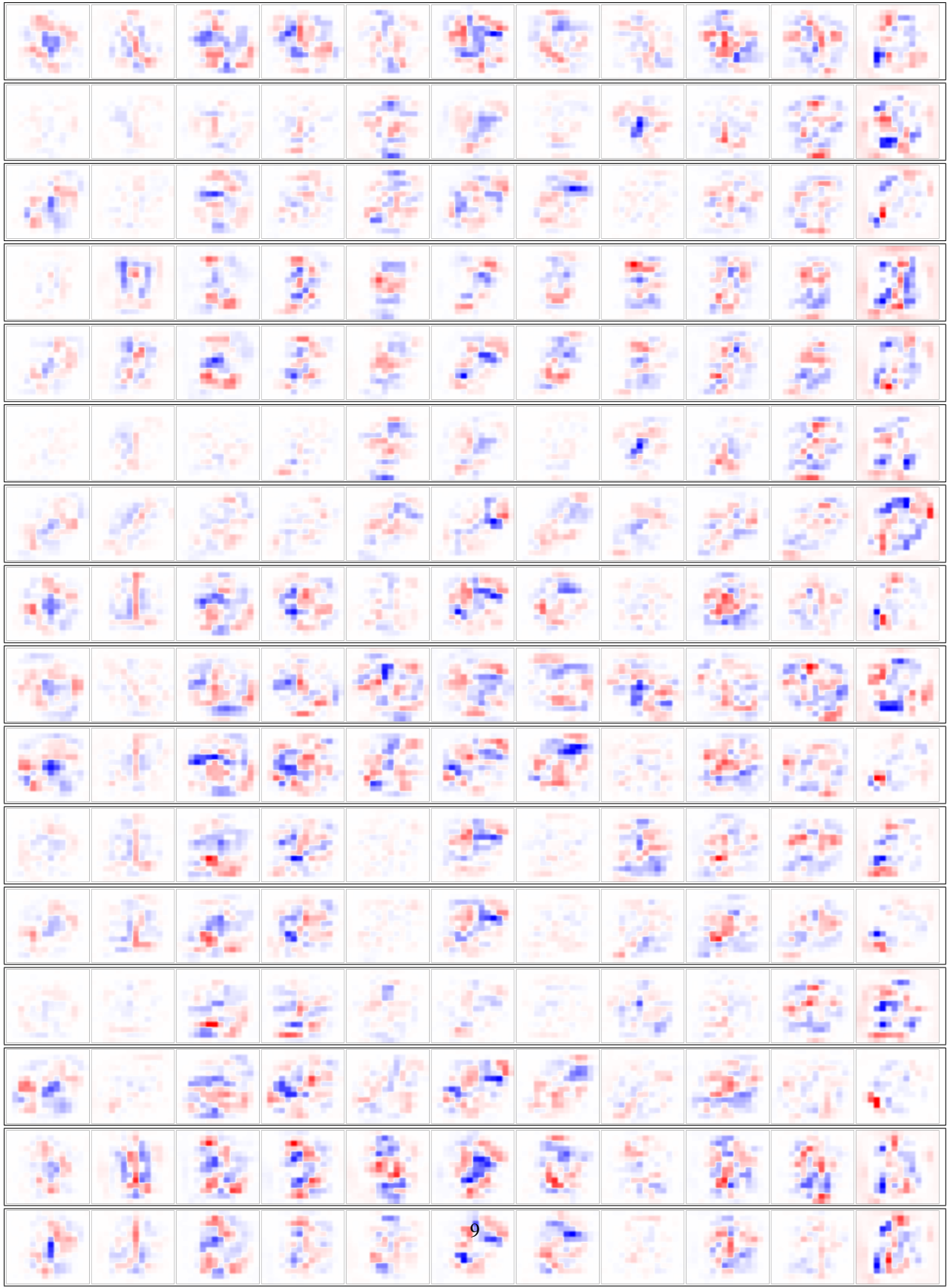

%begin images with
    \caption{All linear weights of IMoIE with $code\_length =4$ trained using all loss terms. Each row corresponds to the linear function of one expert, with columns corresponding to classes 0 through 9 and the rightmost column shows the interpretable routing function of this expert. Red is positive, blue is negative.}
    \label{fig:full_with_reject}
\imageswithrouters
\end{figure}

Figure \ref{fig:full_without_reject} shows all of the learned experts when IMoIE is trained using just $L_{\text{C}}$ and $L_1$. It can be seen that each expert has in fact just learned a constant function. Each expert has very negative weights in the background for one class. Since the background is always valued -1 in MNIST, these blue background weights can be interpreted as a very large positive bias for that class. In other words, each expert is a constant function which always predicts the same class, regardless of input, with different experts corresponding to different classes. The black-box encoder then simply makes the classification decision itself and routes the input to an expert which outputs the desired class. Figure \ref{fig:full_with_reject} shows all of the learned experts when using all the loss terms. It can be seen that the different experts specialize to different variants of digits, for example the first row has specialized to classify digits that are slanted to the left and the fifth row has specialized to digits that are slanted to the right.

\subsection{Effect of Varying the Number of Experts on Accuracy}

\begin{table}
\begin{center}
\begin{tabular}{ |c|c|>{\centering\arraybackslash}p{0.16\linewidth}|>{\centering\arraybackslash}p{0.16\linewidth}|} 
 \hline
 code length & number of experts & accuracy (\%) & explicit accuracy (\%) \\ 
 \hline
 \hline
 4 & 16 & 95.63 & 96.46 \\
 \hline
 8 & 256 & 98.10 & 98.32 \\ 
 \hline
 12 & 4096 & 98.44 & 98.48 \\ 
 \hline
 16 & 65536 & 98.76 & 98.82 \\ 
 \hline
 - & $\infty$ & 99.41 & - \\ 
 \hline
\end{tabular}
\end{center}
\caption{\label{tab:acc} Test accuracy on MNIST10 for varying code lengths. Accuracy refers to the accuracy of the model when the black-box encoder is used to select codes for each input. Explicit accuracy refers to performance in explicit mode with all experts materialized. The final row shows accuracy when the code vector length is set to 256 and the Binarize function is not applied, allowing the code vector to take on continuous values, implicitly defining $\infty$ experts.}
\end{table}

We train several IMoIE with varying code sizes and report the resulting accuracies in Table \ref{tab:acc}. We see that converting the model to explicit mode has almost no effect on accuracy, indicating that the black-box encoder has learned to make interpretable routing decisions. Table \ref{tab:acc} shows that as we increase the number of experts used in the model the accuracy steadily improves. We additionally train a model without using Binarize, effectively defining an IMoIE with an infinite number of experts, and see that this model matches typical black-box classification accuracy.

\subsection{Local Interpretability} 
\begin{figure}[!ht]
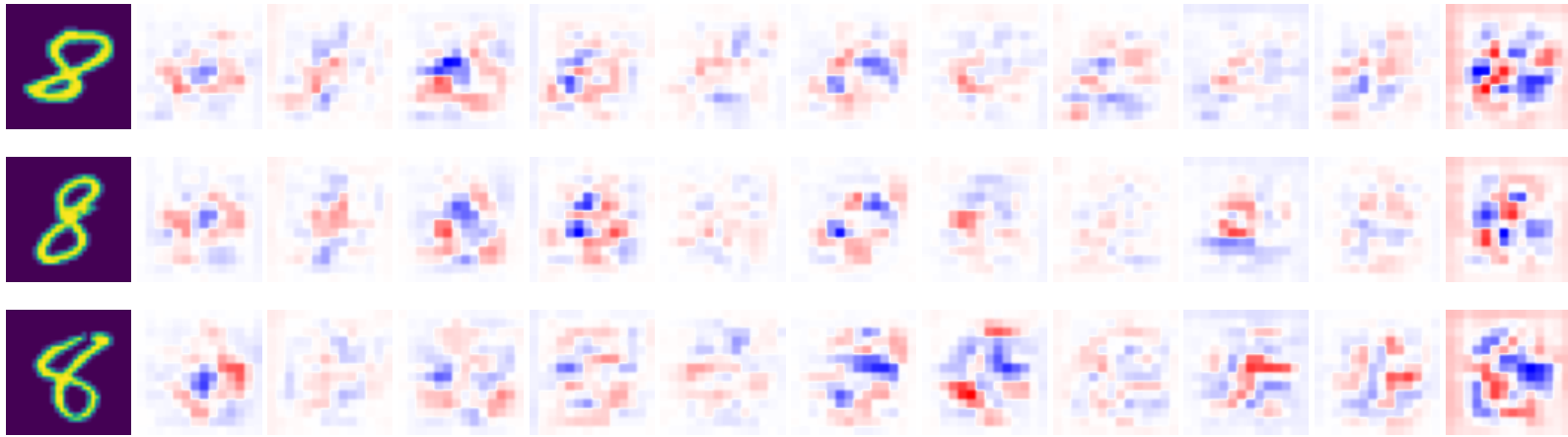

\caption{Various '8' test images and the corresponding experts selected by the black-box router with $code\_length=256$ and no Binarize. Each row corresponds to the linear function of one expert, with columns corresponding to classes 0 through 9 and the rightmost column shows the interpretable routing function of this expert. Red is positive, blue is negative}
\localexplanationinf
\label{fig:localinf}
\end{figure}

When the code length is sufficiently small, IMoIE is globally interpretable, as every expert can be inspected to understand how the model will behave in all situations. Even when the number of experts is prohibitively large, IMoIE can still provide local interpretations by showing the expert that is selected by the black-box encoder for a particular input. Figures \ref{fig:local} and \ref{fig:localinf} show the experts used to classify some test samples for IMoIE with 65536 and $\infty$ experts respectively. It can be seen that the learned experts are very similar for the two models, indicating that even when the number of experts is infinite, each expert will still learn a function that is valid in a small region. Furthermore, these examples show how different inputs are classified using different linear functions, and hence how much each part of the input image contributes towards each class. For example, the images in the column corresponding to class '3' show how the input '8' could be different in order for it to be classified as a '3'. The blue weights in the images show which pixels of the input '8' can be changed from 1 to -1 to turn the input into a class '3' image. It can be seen that the left sides of both of the '8's loops can be removed to change it into a 3.

\begin{figure}[!ht]
\caption{Various '8' test images and the corresponding experts selected by the black-box router with $code\_length=16$. Each row corresponds to the linear function of one expert, with columns corresponding to classes 0 through 9 and the rightmost column shows the interpretable routing function of this expert. Red is positive, blue is negative}
\localexplanation
\label{fig:local}
\end{figure}

\subsection{Ablations}

\begin{table}
\begin{center}
\begin{tabular}{|c|c|c|c|c|c|} 
 \hline
 code length & baseline & $\lambda_{IC}=0$ & $\lambda_{EC}=0$ & $\lambda_{route}=0$ & naive MoIE \\ 
 \hline
 \hline
 4 & 96.46 & 96.21 & 95.68 & 96.43 & 97.56 \\
 \hline
 8 & 98.32 & 98.22 & 96.86 & 97.56 & 97.96\\ 
 \hline
 12 & 98.48 & 98.52 & 96.01 & 97.55 & 98.30 \\ 
 \hline
 16 & 98.82 & 98.63 & 93.31 & 97.74 & 98.19 \\ 
 \hline
\end{tabular}
\end{center}
\caption{\label{tab:ablation} Explicit mode test accuracy (\%) on MNIST10 when different loss terms are removed (weight set to 0). Baseline refers to IMoIE with all loss terms. Naive MoIE refers to a model where all experts and routers are explicitly parameterised, with no encoder or decoder networks.}
\end{table}

To verify the effectiveness of our training methodology we evaluate our model when different components are removed. For each of the 3 loss terms, we train a model with that term's weight set to 0, and evaluate its explicit mode test accuracy. We see that removing any of the loss terms significantly degrades performance. In addition, we train a naive MoIE where all experts and routers are explicitly parameterised as linear functions and trained directly to maximize classification performance. While the naive MoIE performs favourably for small code lengths, when the code length is increased beyond 12 the naive MoIE test accuracy decreases, while its training accuracy is 99.99\%, indicating significant overfitting. On the other hand, IMoIE's performance steadily increases with the number of experts, demonstrating that the convolutional architecture used in the encoder and decoder networks is capable of preventing overfitting in the induced experts.

\section{Discussion}
Traditionally, global and local interpretabillity have been considered as discrete binary possibilities: either a model is globally interpretable or it is locally interpretable. However, our framework suggests that there is a spectrum between the two. Our IMoIE constructs a function by stitching together many interpretable experts. The router can be thought of as indicating for which regions of the input space a particular expert is valid. Thus, by inspecting an expert a human can understand the model's behaviour at all input points where this expert is active. If a human inspects every expert then they will understand the model's behaviour at all possible input points. Thus the degree of global interpretability of a MoIE is directly related to the number of experts it contains. If the MoIE contains a small number of experts, then it is easy to inspect all of them and achieve global understanding. In the limit, with infinitely many experts, each expert is only active for a single infinitesimal input point, and each expert can be considered as a local interpretation. However, there also exists a wide spectrum in between, where there may be too many experts to inspect all of them, but each expert may still be active for a finite region of the input space, and it may still be possible to gain understanding over chunks of the input space by inspecting many experts. Our results show a clear trade off between global interpretability and classification performance: in order to achieve black-box classification accuracy it is necessary to use many experts. Interestingly, our framework does not imply any trade off whatsoever between local interpretability and classification performance.

\section{Conclusion and Future Work}

We have shown that, while the naive implementation of MoE is prone to solve problems solely using the black-box encoder and disregarding the experts, with a suitable training scheme it is possible to create MoIE which provide useful interpretations.

We have developed an implicit parameterisation scheme, whereby experts and routers are generated by a deep neural network, instead of being stored in memory explicitly. This implicit parameterisation enables us to use an arbitrarily large number of experts, and also significantly improves the trainability of MoE models. This implicit parameterisation may be of independent interest outside of the XAI field.

Our IMoIE can be distilled into a fully interpretable model by evaluating in explicit mode, and we demonstrate that the explicit model induced by convolutional neural networks achieves better performance than an explicit model trained directly. In particular, the neural network induced model is significantly more robust to overfitting.

The major drawback of our proposed method is that we require a rather large number of experts before the classification accuracy begins to match that of a black-box classifier. This does not pose a limitation for local interpretability, as our model with infinitely many experts can match black-box classifier accuracy while still providing useful local interpretations. It does, however, severely impede global interpretability. Thus, improving the classification performance of IMoIE while using as few experts as possible is an exciting direction to achieve high performance globally interpretable models. There are numerous ways to improve the proposed IMoIE, some of which include:
\begin{itemize}
\item \textbf{Using more sophisticated functions as experts.} We have so far only used linear functions, which can be thought of as computing the dot product between the input and a learned weight vector. For images, we could instead compute a convolution between inputs and learned filters. This would allow a single expert to match all translations of a particular image, instead of requiring separate experts to match the essentially the same image in different locations.
\item \textbf{Exploring more sophisticated training objectives.} We have so far used a very simple contrastive loss to train the interpretable routers, but this loss may be sub-optimal because it relies on random samples. Similarly, when evaluating the explicit classification loss we only materialize a random sample of all experts due to computational constraints.
\item \textbf{Exploring more sophisticated discretization strategies.} We have employed Binarize with straight-through estimation due to its simplicity, but it is possible that other discretization methods, such as vector quantization \cite{VQ_VAE}, would work better.
\end{itemize}

\bibliographystyle{plain}
\bibliography{refs.bib}

\begin{thebibliography}{10}

\bibitem{xai_survey1}
Amina Adadi and Mohammed Berrada.
\newblock Peeking inside the black-box: a survey on explainable artificial
  intelligence (xai).
\newblock {\em IEEE access}, 6:52138--52160, 2018.

\bibitem{ste}
Yoshua Bengio, Nicholas L{\'e}onard, and Aaron Courville.
\newblock Estimating or propagating gradients through stochastic neurons for
  conditional computation.
\newblock {\em arXiv preprint arXiv:1308.3432}, 2013.

\bibitem{bcos}
Moritz B{\"o}hle, Mario Fritz, and Bernt Schiele.
\newblock B-cos networks: Alignment is all we need for interpretability.
\newblock In {\em Proceedings of the IEEE/CVF Conference on Computer Vision and
  Pattern Recognition}, pages 10329--10338, 2022.

\bibitem{xai_survey3}
Valerie Chen, Jeffrey Li, Joon~Sik Kim, Gregory Plumb, and Ameet Talwalkar.
\newblock Interpretable machine learning: Moving from mythos to diagnostics.
\newblock {\em Queue}, 19(6):28--56, 2022.

\bibitem{MNIST10}
Li~Deng.
\newblock The mnist database of handwritten digit images for machine learning
  research.
\newblock {\em IEEE Signal Processing Magazine}, 29(6):141--142, 2012.

\bibitem{xai_survey2}
Prashant Gohel, Priyanka Singh, and Manoranjan Mohanty.
\newblock Explainable ai: current status and future directions.
\newblock {\em arXiv preprint arXiv:2107.07045}, 2021.

\bibitem{resnet}
Kaiming He, Xiangyu Zhang, Shaoqing Ren, and Jian Sun.
\newblock Deep residual learning for image recognition.
\newblock In {\em Proceedings of the IEEE conference on computer vision and
  pattern recognition}, pages 770--778, 2016.

\bibitem{MoIE}
Aya~Abdelsalam Ismail, Sercan~{\"O} Arik, Jinsung Yoon, Ankur Taly, Soheil
  Feizi, and Tomas Pfister.
\newblock Interpretable mixture of experts for structured data.
\newblock {\em arXiv preprint arXiv:2206.02107}, 2022.

\bibitem{tcav}
Been Kim, Martin Wattenberg, Justin Gilmer, Carrie Cai, James Wexler, Fernanda
  Viegas, et~al.
\newblock Interpretability beyond feature attribution: Quantitative testing
  with concept activation vectors (tcav).
\newblock In {\em International conference on machine learning}, pages
  2668--2677. PMLR, 2018.

\bibitem{adam}
Diederik~P Kingma and Jimmy Ba.
\newblock Adam: A method for stochastic optimization.
\newblock {\em arXiv preprint arXiv:1412.6980}, 2014.

\bibitem{phl1}
Scott~M Lundberg and Su-In Lee.
\newblock A unified approach to interpreting model predictions.
\newblock {\em Advances in neural information processing systems}, 30, 2017.

\bibitem{protoptrees}
Meike Nauta, Ron van Bree, and Christin Seifert.
\newblock Neural prototype trees for interpretable fine-grained image
  recognition.
\newblock In {\em Proceedings of the IEEE/CVF Conference on Computer Vision and
  Pattern Recognition}, pages 14933--14943, 2021.

\bibitem{phl2}
Marco~Tulio Ribeiro, Sameer Singh, and Carlos Guestrin.
\newblock " why should i trust you?" explaining the predictions of any
  classifier.
\newblock In {\em Proceedings of the 22nd ACM SIGKDD international conference
  on knowledge discovery and data mining}, pages 1135--1144, 2016.

\bibitem{scaling_moe}
Carlos Riquelme, Joan Puigcerver, Basil Mustafa, Maxim Neumann, Rodolphe
  Jenatton, Andr{\'e} Susano~Pinto, Daniel Keysers, and Neil Houlsby.
\newblock Scaling vision with sparse mixture of experts.
\newblock {\em Advances in Neural Information Processing Systems},
  34:8583--8595, 2021.

\bibitem{phl3}
Ramprasaath~R Selvaraju, Michael Cogswell, Abhishek Das, Ramakrishna Vedantam,
  Devi Parikh, and Dhruv Batra.
\newblock Grad-cam: Visual explanations from deep networks via gradient-based
  localization.
\newblock In {\em Proceedings of the IEEE international conference on computer
  vision}, pages 618--626, 2017.

\bibitem{VQ_VAE}
Aaron Van Den~Oord, Oriol Vinyals, et~al.
\newblock Neural discrete representation learning.
\newblock {\em Advances in neural information processing systems}, 30, 2017.

\bibitem{xai_survey4}
Yu~Zhang, Peter Ti{\v{n}}o, Ale{\v{s}} Leonardis, and Ke~Tang.
\newblock A survey on neural network interpretability.
\newblock {\em IEEE Transactions on Emerging Topics in Computational
  Intelligence}, 2021.

\end{thebibliography}

\end{document}